%% file: main.tex
\documentclass[10pt,twocolumn,letterpaper]{article}

\usepackage[pagenumbers]{cvpr} 
\usepackage{graphicx}
\usepackage{amsmath}
\usepackage{amssymb}
\usepackage{booktabs}
\usepackage{cuted}
\usepackage{multirow}
\usepackage[accsupp]{axessibility}
\usepackage[pagebackref,breaklinks,colorlinks]{hyperref}
\usepackage[usenames, dvipsnames]{color}
\newcommand\norm[1]{\left\lVert#1\right\rVert}

\newcommand{\ourmethod}[1]{RAC}

\usepackage[capitalize]{cleveref}
\crefname{section}{Sec.}{Secs.}
\Crefname{section}{Section}{Sections}
\Crefname{table}{Table}{Tables}
\crefname{table}{Tab.}{Tabs.}

\def\cvprPaperID{1184} %
\def\confName{CVPR}
\def\confYear{2023}

\author{Gengshan Yang \; Chaoyang Wang \; N Dinesh Reddy \; Deva Ramanan\\[3pt]
Carnegie Mellon University\\
}

\begin{document}

\title{Reconstructing Animatable Categories from Videos}

\maketitle
\begin{strip}\centering
\vspace{-30pt}
    \includegraphics[width=\linewidth, trim={1.8cm 8cm 1.8cm 8.5cm},clip]{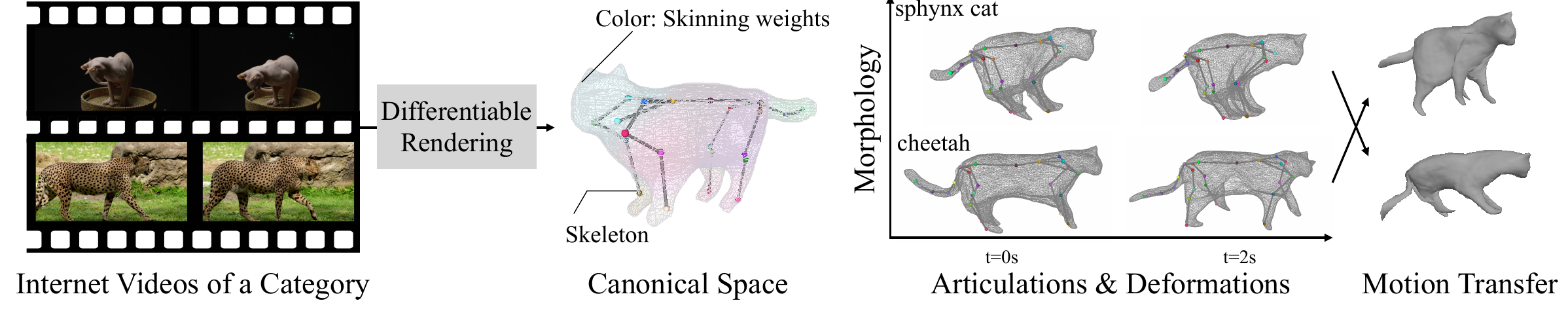}
\captionof{figure}{Given videos of a deformable category and a skeleton, we reconstruct an animatable 3D model that factorizes variations {\em across} instances (e.g., {\tt cheetah}'s and {\tt sphynx}'s are both {\tt cat}s but with different shape morphology, skeleton dimensions, and texture)
from time-specific variations {\em within} an instance (e.g., skeleton articulations and elastic shape deformation). \textbf{Left}: Input videos; \textbf{Middle-left}: 3D shape, skeleton, and skinning weights (visualized as surface colors) in the canonical space; \textbf{Middle-right}:
Disentangled between-instance and within-instance variations over time. \textbf{Right}:
Morphology and motion transferred across the two instances.
\label{fig:teaser}}
\end{strip}

\begin{abstract}

Building animatable 3D models is challenging due to the need for 3D scans, laborious registration, and manual rigging, which are difficult to scale to arbitrary categories.
Recently, differentiable rendering provides a pathway to obtain high-quality 3D models from monocular videos, but these are limited to rigid categories or single instances. 
We present RAC that builds category 3D models from monocular videos while disentangling variations over instances and motion over time. Three key ideas are introduced to solve this problem: (1) specializing a skeleton to instances via optimization, (2) a method for latent space regularization that encourages shared structure across a category while maintaining instance details, and (3) using 3D background models to disentangle objects from the background. We show that 3D models of humans, cats and dogs can be learned from 50-100 internet videos. Project page: \href{https://gengshan-y.github.io/rac-www/}{https://gengshan-y.github.io/rac-www/}.

\end{abstract}

\section{Introduction}
\label{sec:intro}

We aim to build animatable 3D models for deformable object categories. Prior work has done so for targeted categories such as people (e.g., SMPL~\cite{SMPL:2015, anguelov2005scape}) and quadruped animals (e.g., SMAL~\cite{biggs2018creatures}), but such methods appear challenging to scale due to the need of 3D supervision and registration. Recently, test-time optimization through differentiable rendering~\cite{pumarola2020d, park2021hypernerf, wang2021nerf, park2021nerfies, yang2022banmo} provides a pathway to generate high-quality 3D models of deformable objects and scenes from monocular videos. However, such models are typically built {\em independently} for each object instance or scene. In contrast, we would like to build \emph{category} models that can generate different instances along with deformations, given {\em causally-captured video collections}. 

\begin{figure*}
    \centering
    \includegraphics[width=\linewidth, trim={3.5cm 8.5cm 3.5cm 8cm},clip]{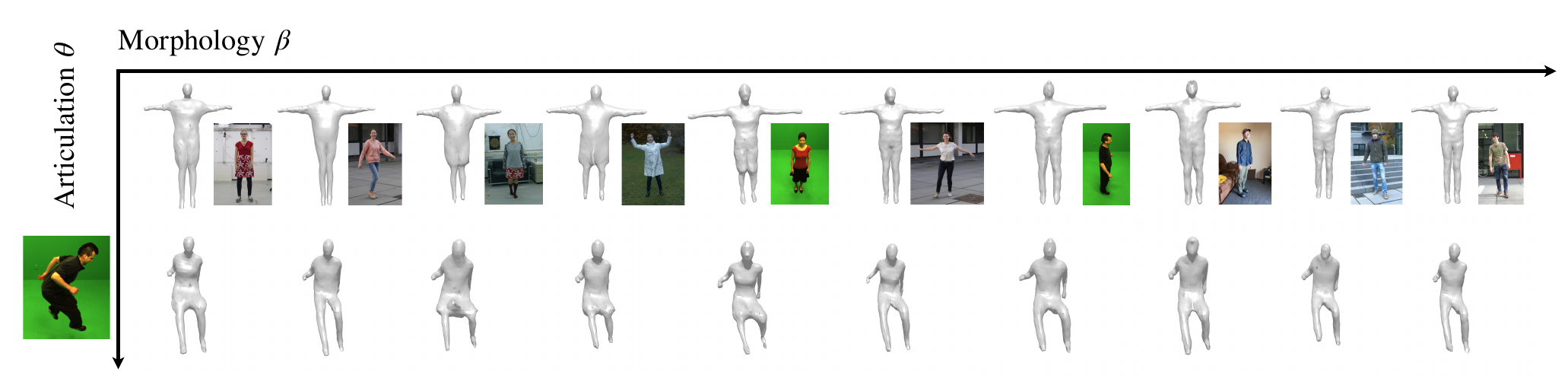}
    \caption{\textbf{Disentangling morphologies $\beta$ and articulation $\theta$.} We show different morphologies (body shape and clothing) given the same rest pose ({\bf top}) and bouncing pose ({\bf bottom}).}
    \label{fig:disentangle}
    \vspace{-10pt}
\end{figure*}

Though scalable, such data is challenging to leverage in practice. One challenge is how to learn the \emph{morphological variation} of instances within a category. For example, {\tt husky}s and {\tt corgi}s are both {\tt dog}s, but have different body shapes, skeleton dimensions, and texture appearance. Such variations are difficult to disentangle from the variations \emph{within} a single instance, e.g., as a dog articulates, stretches its muscles, and even moves into different illumination conditions. Approaches for disentangling such factors require enormous efforts in capture and registration~\cite{anguelov2005scape, bogo2014faust}, and doing so without explicit supervision remains an open challenge.

Another challenge arises from the impoverished nature of in-the-wild videos: objects are often \emph{partially observable} at a limited number of viewpoints, and input signals such as segmentation masks can be inaccurate for such ``in-the-wild" data. When dealing with partial or impoverished video inputs, one would want the model to listen to the common structures learned across a category -- e.g., dogs have two ears. On the other hand, one would want the model to stay faithful to the input views.

Our approach addresses these challenges by exploiting three insights: 
(1) We learn skeletons with constant bone lengths within a video, allowing for better disentanglement of between-instance morphology and within-instance articulation. (2) We regularize the unobserved body parts to be coherent across instances while remaining faithful to the input views with a novel code-swapping technique. (3) We make use of a category-level background model that, while not 3D accurate, produces far better segmentation masks. 
We learn animatable 3D models of cats, dogs, and humans which outperform prior art. Because our models register different instances with a canonical skeleton, we also demonstrate motion transfer across instances.

\section{Related Works}

\noindent\textbf{Model-based 3D Reconstruction.} 
A large body of work in 3D human and animal reconstruction uses parametric shape models~\cite{SMPL:2015,SMPL-X:2019,xiang2019monocular,vo2020spatiotemporal,Zuffi:CVPR:2018,Zuffi:CVPR:2017}, which are built from registered 3D scans of human or animals, and serve to recover 3D shapes given a single image or video at test time ~\cite{badger2020,biggs2020wldo,kocabas2019vibe,zuffi2019three,kocabas2019vibe, BARC:2022, biggs2020wldo}.  A recent research focus is to combine statistical human body mode with implicit functions~\cite{saito2019pifu,saito2020pifuhd,xiu2022icon, zheng2021pamir, Saito:CVPR:2021, li2020monoportRTL, jiang2022selfrecon} to improve the robustness and fidelity of the reconstruction. Although parametric body models achieve great success in reconstructing humans with large amounts of ground-truth 3D data, it is unclear how to apply the same methodology to categories with limited 3D data, such as animals, and how to scale to real-life imagery with diverse clothing and body poses. \ourmethod{} builds category-level shape models from in-the-wild videos and demonstrates the potential to reconstruct 3D categories without sophisticated manual processing.

\noindent\textbf{Category Reconstruction from Image Collections.} 
A number of recent methods build deformable 3D models of object categories from images with weak 2D annotations, such as keypoints and object silhouettes, obtained from human annotators or predicted by off-the-shelf models~\cite{Ye_2021_CVPR,ucmrGoel20,cmrKanazawa18,li2020self, tulsiani2020imr, kokkinos2021point, wu2022magicpony}. However, those methods do not distinguish between morphological variations and motion over time. Moreover, they often apply heavy regularization on shape and deformation to avoid degenerate solutions, which also smooths out fine-grained details. Recent research combines neural implicit functions~\cite{mildenhall2020nerf, mescheder2019occupancy} with category modeling in the context of 3D data generation~\cite{niemeyer2021giraffe, chanmonteiro2020pi-GAN,Chan2021}, where shape and appearance variations over a category are modeled with conditional NeRFs. However, reconstructions are typically focused on near-rigid objects such as faces and vehicles. 

\noindent\textbf{Articulated Object Reconstruction from Videos.} 
Compared to image collections, videos provide signals to reconstruct object motions and disentangle them from morphological variations. Some works~\cite{li2022tava, su2021anerf, 2021narf} reconstruct articulated bodies from videos, but they either assume synchronized multi-view recordings or articulated 3D skeleton inputs that make their approaches less general. Some other works~\cite{yang2022banmo, yang2021lasr, yang2021viser} learn animatable 3D models from monocular videos capturing the same object instance, without disentangling morphology and motion. There are recent methods~\cite{vmr2020, wu2021dove, yang2021viser} using in-the-wild videos to reconstruct 3D models animals, but their quality are relatively low. 

\begin{figure*}
    \centering
    \includegraphics[width=\linewidth, trim={0.5cm 8.7cm 0.5cm 8.5cm},clip]{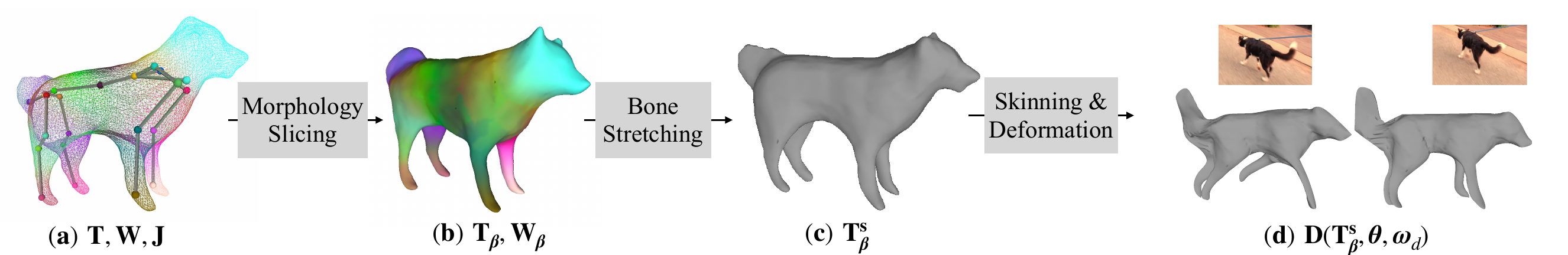}
    \caption{\textbf{Morphological variations vs time-varying articulation and deformation.} (a) Canonical shape ${\bf T}$, skinning weights ${\bf W}$, and joint locations ${\bf J}$. (b) To represent morphological differences between instances, we use a morphology code $\boldsymbol\beta$ that specifies instance shape and appearance ${\bf T}_{\boldsymbol\beta}$, skinning weights ${\bf W}_{\boldsymbol\beta}$ for a canonical skeleton ${\bf J}$. (c) $\boldsymbol\beta$ also predicts a change in bone lengths $\Delta{\bf J}_{\boldsymbol\beta}$ which further {\em stretches} instance shape into ${\bf T}^s_{\boldsymbol\beta}$ by elongating body parts. 
    (d) Time-varying articulations are modeled with an articulation vector ${\boldsymbol\theta}$ by linearly blending rigid bone transformations in the dual quaternion space. Time-varying deformations (such as muscle deformation) are modeled with a deformation vector ${\boldsymbol\omega_d}$ through invertible 3D warping fields.
    }
    \label{fig:model}
    \vspace{-10pt}
\end{figure*}

\section{Method}
Given video recordings of different instances from a category and a pre-defined skeleton, we build animatable 3D models including instance-specific morphology (Sec.~\ref{sec: morphology}), time-varying articulation and deformation (Sec.~\ref{sec: motion}), as well as a video-specific 3D background model (Sec.~\ref{sec:background}). The models are optimized using differentiable rendering (Sec.~\ref{sec:opt}). An overview is shown in Fig.~\ref{fig:model}.

\subsection{Between-Instance Variation}\label{sec: morphology}
Fusing videos of different instances into a category model requires handling the morphological variations, which includes the changes in both \emph{internal skeleton} and \emph{outward appearance} (shape and color). We define a video-specific morphology code $\boldsymbol\beta$ to control the variations of both the shape and the skeleton. 

To model between-instance shape variations, one could use dense warping fields to deform a canonical template into instance-specific shapes~\cite{xie2021fig}. However, warping fields cannot explain topological changes (e.g., different clothing). Instead, we define a hierarchical representation: a conditional canonical field~\cite{weng_humannerf_2022_cvpr, park2021hypernerf, chan2021pi} to handle fine-grained variations over a category (e.g., the ears of dogs) and a stretchable bone model~\cite{jacobson2011stretchable, wu2022casa} to represent coarse shape variations (e.g., height and size of body parts). 

\noindent{\bf Conditional Field ${\bf T}$.}
In the canonical space, a 3D point $\mathbf{X} \in \mathbb{R}^3$ is associated with three properties: signed distance $d \in \mathbb{R}$, color $\mathbf{c} \in \mathbb{R}^3$, and canonical features $\boldsymbol{\psi} \in \mathbb{R}^{16}$, which is used to register pixel observations to the canonical space~\cite{Neverova2020cse, yang2021viser}.
These properties are predicted by multi-layer perceptron (MLP) networks: 
\begin{align}
    (d,{\bf c}^t) &= \mathbf{MLP}_\mathrm{SDF}({\bf X},\boldsymbol{\beta},\boldsymbol{\omega}_{a}), \label{eq:density}\\
    \boldsymbol{\psi} &= \mathbf{MLP}_{\boldsymbol\psi}({\bf X}) \label{eq:feature},
\end{align}
where the shape and color are conditioned on a video-specific morphology code $\boldsymbol{\beta}{\in}\mathbb{R}^{32}$~\cite{jang2021codenerf, niemeyer2021giraffe}. We further ask the color to be dependent on an appearance code $\boldsymbol{\omega}_{a}{\in}\mathbb{R}^{64}$ that captures frame-specific appearance such as shadows and illumination changes~\cite{martinbrualla2020nerfw}. 

\noindent{\bf Skeleton $\bf{J}$.}
Unlike shape and color, the bone structures are not directly observable from imagery, making it ambiguous to infer. Methods for automatic skeletal rigging~\cite{le2014robust,noguchi2021watch, RigNet} either heavily rely on shape priors, or appear sensitive to input data. Instead, we provide a category-level skeleton topology, which has a fixed tree topology with (B+1) bones and B joints (B=25 for quadruped and B=18 for human). To model cross-instance morphological changes, we define per-instance joint locations as:
\begin{align}
{\bf J} & = \mathbf{MLP}_{\bf J}(\boldsymbol\beta)\in\mathbb{R}^{3\times B}.
\end{align}
As we will discuss next, the change in joint locations not only stretches the skeleton, but also results in the elongation of canonical shapes as shown in Fig.~\ref{fig:model} (c). The skeleton topology is fixed through optimization but ${\bf J}$ is specialized to each video.

\noindent{\bf Skinning Field $\bf{W}$.}
For a given 3D location ${\bf X}$, we define skinning weight ${\bf W} \in \mathbb{R}^{B+1}$ following BANMo:
\begin{align}
      {\bf W} & = \sigma_{\mathrm{softmax}}\bigl({d}_\sigma({\bf X},\boldsymbol\beta,\boldsymbol{\theta})+\mathbf{MLP}_{\bf W}({\bf X},\boldsymbol\beta,\boldsymbol{\theta})\bigr),\label{eq:skinning}
\end{align}
where $\boldsymbol\theta$ is a articulation code and $d_{\sigma}({\bf X},\boldsymbol\beta, \boldsymbol{\theta})$ is the Mahalanobis distance between ${\bf X}$ and Gaussian bones under articulation $\boldsymbol\theta$ and morphology $\boldsymbol\beta$, refined by a delta skinning MLP. Each Gaussian bone has three parameters for center, orientation, and scale respectively, where the centers are computed as the midpoint of two adjacent joints, the orientations are determined by the parent joints, and the scales are optimized. 

\noindent\textbf{Stretchable Bone Deformation.} To represent variations of body dimension and part size, prior work~\cite{SMPL:2015,Zuffi:CVPR:2017} learns a PCA basis from registered 3D scans. Since 3D registrations are not available for in-the-wild videos, we optimize a parametric model through differentiable rendering. Given the stretched joint locations, the model deforms the canonical shape ${\bf T}_{\boldsymbol\beta}$ with blend skinning equations,
\begin{equation}\label{eq:stretch}
{\bf T}^s_{\boldsymbol\beta}= \big({\bf W}_{\boldsymbol\beta}{\bf G_{\boldsymbol\beta}}\big) {\bf T}_{\boldsymbol\beta},
\end{equation}
where ${\bf G}_{\boldsymbol\beta}$ transforms the bone coordinates, and ${\bf W}_{\boldsymbol\beta}$ is the instance-specific skinning weights in Eq.~\eqref{eq:skinning}. 

\subsection{Within-Instance Variation}\label{sec: motion}
We represent within-instance variations as time-varying warp fields between the canonical space and posed space at time $t$. Similar to HumanNeRF~\cite{weng_humannerf_2022_cvpr}, we decompose motion as \emph{articulations} that explains the near-rigid component (e.g., skeletal motion) and \emph{deformation} that explains the remaining nonrigid movements (e.g., cloth deformation). Note given the complexity of body movements, it is almost certain the pre-defined skeleton would ignore certain movable body parts. Adding deformation is crucial to achieving high-fidelity reconstruction.

\begin{figure*}
    \centering
    \includegraphics[width=\linewidth, trim={2.5cm 9cm 2.5cm 9cm},clip]{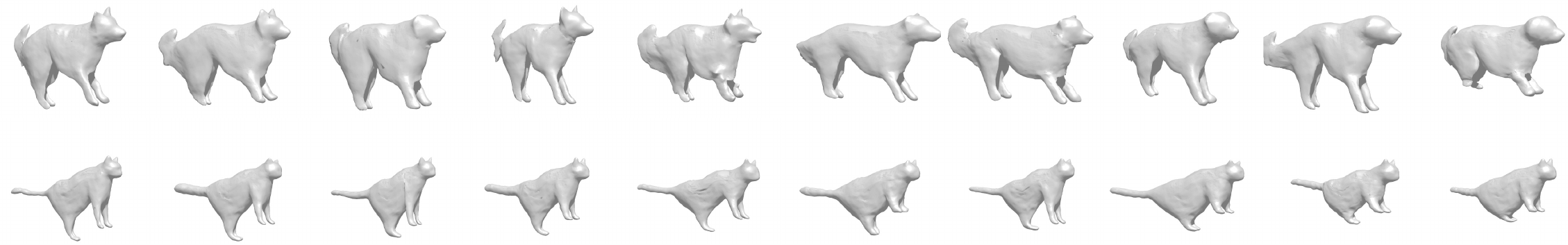}
    \caption{\textbf{Different $\boldsymbol\beta$ morphologies of dogs (top) and cats (bottom)}. Our reconstructions show variance in ear shape and limb size over dog breeds, as well as variance in limb and body size over cat breeds.}
    \label{fig:gallery}
    \vspace{-10pt}
\end{figure*}

\noindent\textbf{Time-varying Articulation.}
To model the time-varying skeletal movements, we define per-frame joint angles:
\begin{equation}
{\bf Q} = \mathbf{MLP}_{\bf A}(\boldsymbol\theta)\in\mathbb{R}^{3\times B}
\end{equation}
where $\boldsymbol\theta\in\mathbb{R}^{16}$ is a low-dimensional articulation parameter, and each joint has three degrees of freedom. Given joint angles and the per-video joint locations, we compute bone transformations ${\bf G}\in\mathbb{R}^{3\times4\times B}$ via forward kinematics. 
We apply dual quaternion blend skinning (DQS)~\cite{kavan2007skinning} to get the warping field for each spatial point,
\begin{equation}\label{eq:articulation}
{\bf D}(\boldsymbol\beta, \theta)=({\bf W}_{\boldsymbol\beta}{\bf G}) {\bf T}^s_{\boldsymbol\beta}.
\end{equation}
Compared to linear blend skinning (LBS), DQS blends SE(3) transformations in dual quaternion space and ensures valid SE(3) after blending, which reduces artifacts around twisted body parts. For more analysis and comparisons, we refer readers to a concurrent work~\cite{song2023moda} that also applies DQS for deformable object reconstruction. Note the stretching operation in Eq.~\eqref{eq:stretch} can be fused with articulation as a single blend skinning operation. 

\noindent\textbf{Time-varying Soft Deformation.} 
To further explain the dynamics induced by non-skeleton movements (such as the cat belly and human clothes), we add a neural deformation field~\cite{park2021nerfies,Lei2022CaDeX} $\mathcal{D}(\cdot)$ that is flexible enough to model highly nonrigid deformation. Applying the fine-grained warping after blend skinning, we have \begin{equation}\label{eq:deformation}
{\bf D}(\boldsymbol\beta, \theta, \omega_d)= {\mathcal D}({{\bf D}(\boldsymbol\beta, \theta)}, \omega_d), 
\end{equation}
where $\omega_d$ is a frame-specific deformation code. Inspired by CaDeX~\cite{Lei2022CaDeX}, we use real-NVP~\cite{dinh2016density} to ensure the 3D deformation fields are invertible by construction.

\noindent\textbf{Invertibility of 3D Warping Fields.} For a given time instance $t$, we have defined a forward warping field $\mathcal{W}^{t,\rightarrow}$ that transforms 3D points from the canonical space to the specified time instance, and a backward warping field $\mathcal{W}^{t,\leftarrow}$ to transform points in the inverse direction. 
Both warping fields include stretching (Eq.~\eqref{eq:stretch}), articulation. (Eq.~\eqref{eq:articulation}), and deformation (Eq.~\eqref{eq:deformation}) operations. Notably, we only need to define each operation in the forward warping fields. The deformation operation is, by construction, invertible. To invert stretching and articulation, we invert SE (3) transformations ${\bf G}$ in the blend skinning equations and compute the skinning weights with Eq.~\eqref{eq:skinning} using the corresponding morphology and articulation codes. A 3D cycle loss is used to ensure that the warping fields are self-consistent after a forward-backward cycle~\cite{yang2022banmo, li2021neural}.

\subsection{Scene Model}\label{sec:background} 

Reconstructing objects from in-the-wild video footage is challenging due to failures in segmentation, which is often caused by out-of-frame body parts, occlusions, and challenging lighting. Inspired by background subtraction~\cite{jain1979analysis, sheikh2009background}, we build a model of the background to \emph{robustify} our method against inaccurate object segmentation. 

\begin{figure}[t!]
    \centering
    \includegraphics[width=\linewidth, trim={10.2cm 7.6cm 10.2cm 7.2cm},clip]{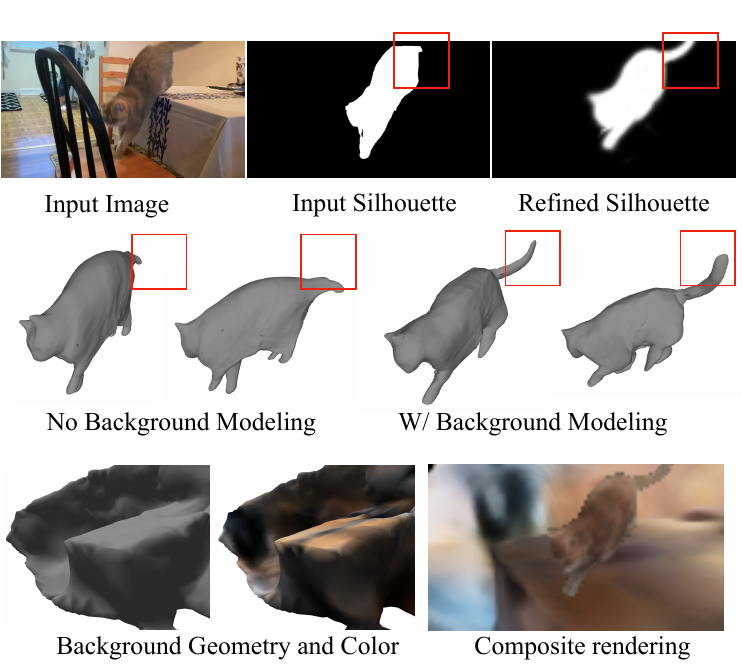}
    \caption{\textbf{Joint foreground and background reconstruction.} We jointly reconstruct objects and their background, while refining the segmentation. Note the input silhouette is noisy (e.g., tail was not segmented), and background modeling helps produce an accurate refined silhouette. As a result, \ourmethod{} is robust to inaccurate segmentation (e.g., tail movements marked by the red box).}
    \label{fig:results-seg}
\end{figure}

In background subtraction, moving objects can be segmented by comparing input images to a background model (e.g., a median image). We generalize this idea to model the background scene in 3D as a per-video NeRF, which can be rendered as color pixels at a moving camera frame and compared to the input frame. We design a conditional background model that generates density and color of a scene conditioned on a per-video background code $\boldsymbol{\gamma}$:
\begin{align}
    (\sigma,{\bf c}^t) &= \mathbf{MLP}_\mathbf{bg}({\bf X},{\bf v}, \boldsymbol{\gamma}), \label{eq:bg}
\end{align}
where ${\bf v}$ is the viewing direction. To render images, we compose the density and color of the object field and the background NeRF in the view space~\cite{niemeyer2021giraffe}, and compute the expected color and optical flow. Background modeling and composite rendering allows us to remove the object silhouette loss, and improves the quality of results. Interestingly, we find that even {\em coarse} geometric reconstructions of the background still can improve the rendered 2D object silhouette, which in turn is useful for improving the quality of object reconstructions (Fig.~\ref{fig:results-seg}). We ablate the design choice in Tab.~\ref{tab:quan-human}.

\subsection{Losses and Regularization} \label{sec:opt}
Given the videos and a predefined skeleton, we optimize the parameters discussed above: (1) canonical parameters $\{\bf \boldsymbol\beta, J,T,W\}$ including per-video morphology codes and canonical templates; (2) motion parameters $\{\bf \boldsymbol\theta, \boldsymbol\omega_d, A,D\}$ including per-frame codes as well as articulation and soft deformation MLPs. (3) background parameters $\{\bf \boldsymbol\gamma, B\}$ including video background codes and a background NeRF. 
The overall objective function contains a image reconstruction loss term and regularization terms.

\noindent{\bf Reconstruction Losses.} The reconstruction loss is defined as the difference between rendered and observed images, including object silhouette, color, flow, and features:
\begin{align}
    \mathcal{L} = \mathcal{L}_{\textrm{sil}} + \mathcal{L}_{\textrm{rgb}} + \mathcal{L}_{\textrm{OF}} + \mathcal{L}_{\textrm{feat}}.
\end{align}
We update the model parameter by minimizing $\mathcal{L}$ through differentiable volume rendering in the same way as BANMo~\cite{yang2022banmo}. Off-the-shelf estimates of object silhouettes are used as supervision to kick-start the optimization. Then the weight of silhouette term is set to 0 after several iterations of optimization, while composite rendering of foreground and background itself is capable of separating the object and the non-object components.

\noindent{\bf Morphology Code ($\boldsymbol\beta$) Regularization.} Existing differentiable rendering methods are able to faithfully reconstruct the input view but not able to hallucinate a reasonable solution for occluded body parts~\cite{yang2022banmo, park2021nerfies}. See Fig.~\ref{fig:results-beta-reg} for an example. One solution is to regularize the instance-specific morphology code $\boldsymbol\beta$ to be consistent with the body shapes observed in {\em other} videos. Traditional approaches might do this by adding variational noise (as in a VAE) or adversarial losses (as in a GAN). We found good results with the following easy-to-implement approach: we randomly {\em swap} the morphology code $\boldsymbol\beta$ of two videos during optimization; this regularizes the model to effectively learn a single morphology code that works well across all videos. But naively applying this approach would produce a single morphology that would not specialize to each object instance. To enable specialization, we {\em gradually} decrease the probability of swaps during the optimization, from  $\mathcal{P}=1.0 \rightarrow 0.05$.

\noindent{\bf Joint ${\bf J}$ Regularization.} Due to the non-observable nature of the the joint locations, there might exist multiple joint configurations leading to similar reconstruction error. To register the skeleton with the canonical shape, we minimize Sinkhorn divergence~\cite{feydy2019interpolating} between the canonical surface ${\bf T}_{\boldsymbol\beta}$ and the joint locations ${\bf J}_{\boldsymbol\beta}$, which forces them to occupy the same space. We extract the canonical mesh with marching cubes~\cite{lorensen1987marching} as a proxy of the canonical surface. Sinkhorn distance interpolates between Wasserstein and kernel distances and defines a soft way to measure the distance between shapes with different density distributions. 

\noindent{\bf Soft Deformation Regularization} The soft deformation field has the capacity of explaining not only the soft deformations, but also the skeleton articulations. Therefore, we penalize the L2 norm of the soft deformation vectors at randomly sampled morphology and articulations,
\begin{equation}\label{eq:stretch-soft}
\mathcal{L}_{\textrm{soft}} = \norm{{\bf D}(\boldsymbol\beta, \theta, \omega_d) - {\bf D}(\boldsymbol\beta, \theta)}.
\end{equation}

\begin{table}
    \caption{\textbf{Quantitative results on \texttt{AMA} sequences.} 3D Chamfer distance (cm, $\downarrow$) and F-score (\%, $\uparrow$) averaged over all frames. Our model is trained on 47 human videos spanning existing human datasets (as described in Sec.4.2); we also train BANMo on the same set. Other baselines are trained on 3D human data and relies on SMPL model. Results with $^S$ indicates variants trained on single instances. Our model outperforms prior works.}
    \small
    \centering
    \resizebox{\columnwidth}{!}{%
    \begin{tabular}{lcccccccccccc}
	\toprule
	\multirow{2.5}{*}{Method}
    &\multicolumn{3}{c}{\texttt{samba}}&\multicolumn{3}{c}{\texttt{bouncing}}\\
\cmidrule(lr){2-4}\cmidrule(lr){5-7}&CD &F@2\%&F@5\%&CD &F@2\%&F@5\\
\midrule
HuMoR &9.8 & 47.5 & 83.7 &11.5 & 45.2 & 82.3\\
ICON &10.1 & 39.9 & 85.2 &9.7 & 53.5 & 86.4\\
BANMo$^{S}$ &8.0 & 62.2& 89.1 & 7.6 & 64.7 & 91.1\\
BANMo & 9.3 & 54.4 & 85.5 & 10.2  & 54.4 & 86.5\\
\ourmethod{}$^{S}$ &6.4 & 70.9 &93.2 & {\bf 6.9} & {\bf 66.7} & {\bf 92.8}\\
\ourmethod{} & {\bf 6.0} & {\bf 72.5} & {\bf 94.4} & 8.0 & 63.8 & 91.4\\
\midrule
w/o skeleton &8.6 & 59.6 & 87.7 &9.3 & 59.5 & 87.8\\ 
w/o $\boldsymbol\beta$ & 8.5 & 58.9 & 87.5 & 8.4 & 62.5 & 90.6 \\
$\boldsymbol\beta$ swap$\rightarrow||\boldsymbol\beta||^2_2$ & 6.5 & 69.0 & 93.8 &8.0& 64.8 & 91.3\\
+ bkgd NeRF & 6.3 & 70.9 & 93.7 &7.4 & 65.5 & 91.8\\
\bottomrule
\label{tab:quan-human}
\end{tabular}
}
\vspace{-10pt}
\end{table}

\begin{figure*}[ht!]
    \centering
    \includegraphics[width=\linewidth, trim={5cm 7.5cm 5.5cm 8cm},clip]{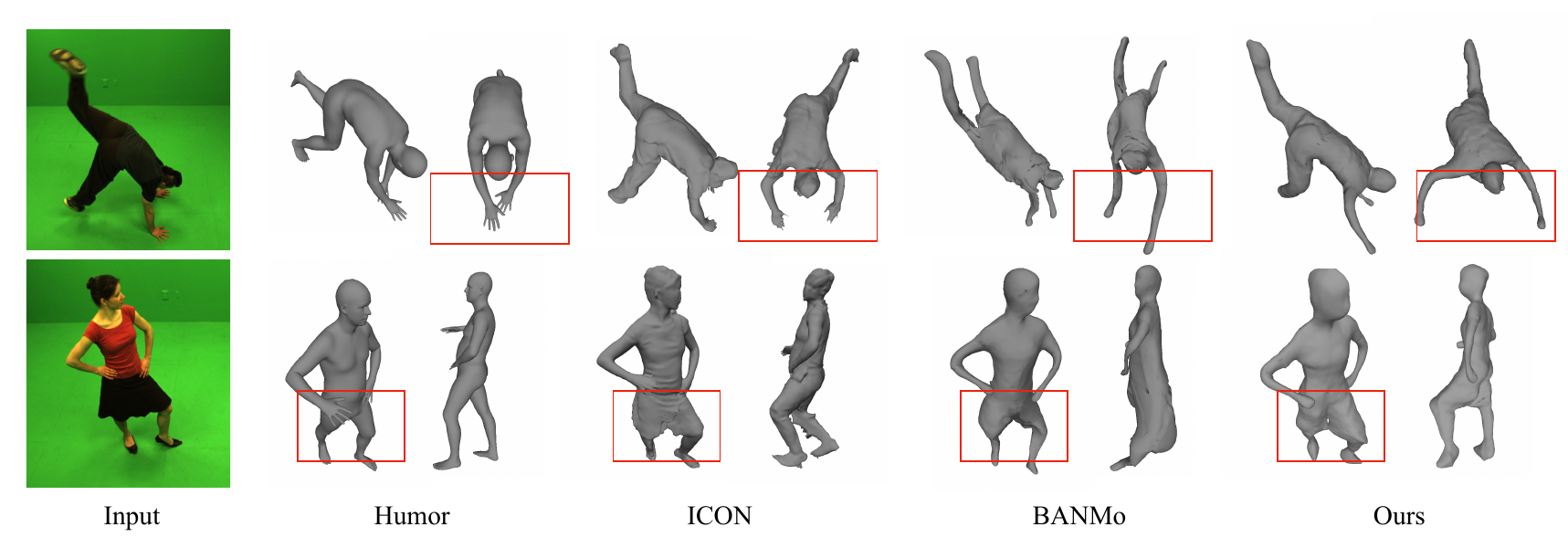}
    \includegraphics[width=\linewidth, trim={0cm 8.5cm 0cm 8cm},clip]{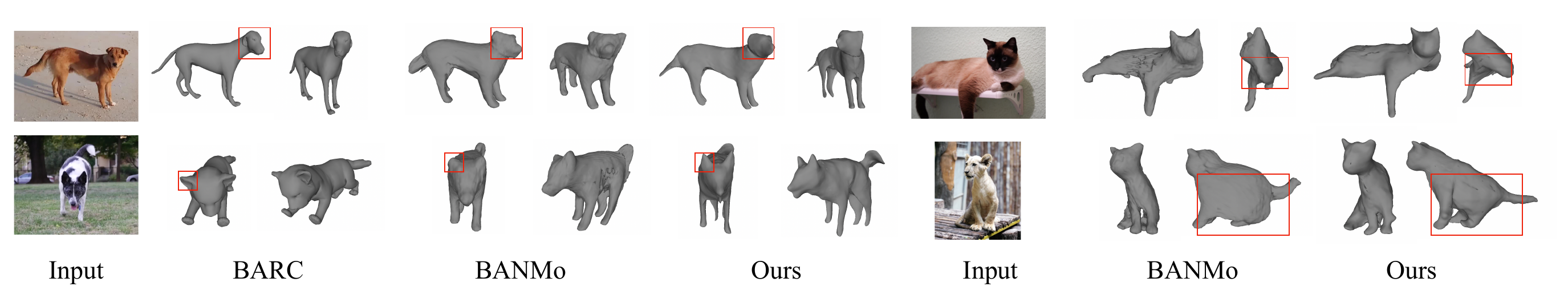}
    \caption{\textbf{Qualitative comparison.} We compare with BANMo and model-based methods (HuMoR, ICON, BARC). {\bf Top}: human reconstruction on (\texttt{AMA}). {\bf Bottom:} dogs and cats reconstruction on internet videos.}
    \label{fig:results-comp}
    \vspace{-10pt}
\end{figure*}

\section{Experiments}
\noindent{\bf Implementation Details.}
We build \ourmethod{} on BANMo and compute bone transformations from a kinematic tree. The soft deformation field follows CaDeX, where we find that two invertible blocks are capable of handling moderate deformations. To evaluate surface reconstruction accuracy, we extract the canonical mesh ${\bf T}$ by finding the zero-level set of SDF with marching cubes on a $256^3$ grid. To get the shape at a specific time, the canonical mesh is forward-warped with $\mathcal{W}^{t,\rightarrow}$. 

\begin{figure}[h!]
    \centering
    \includegraphics[width=\linewidth, trim={3cm 8.5cm 3cm 8.5cm},clip]{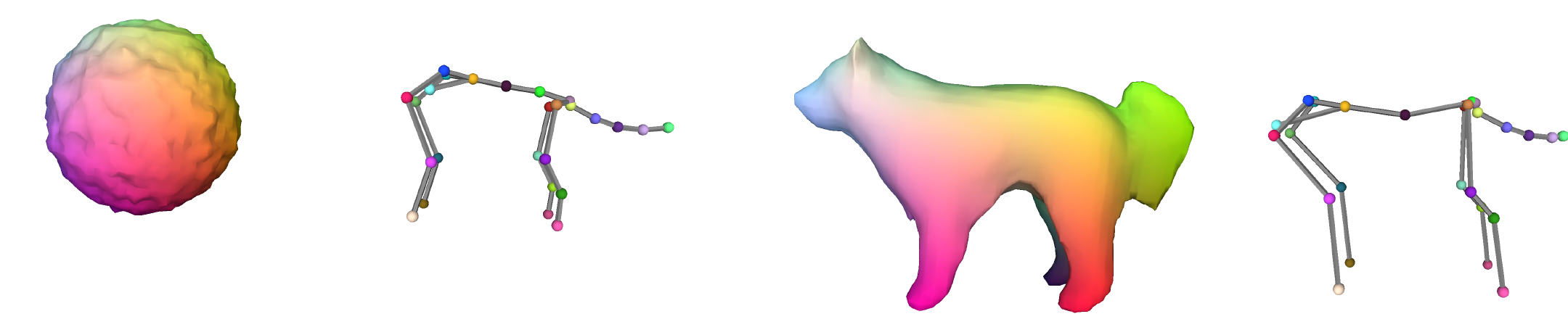}
    \includegraphics[width=\linewidth, trim={3cm 8.5cm 3cm 8.5cm},clip]{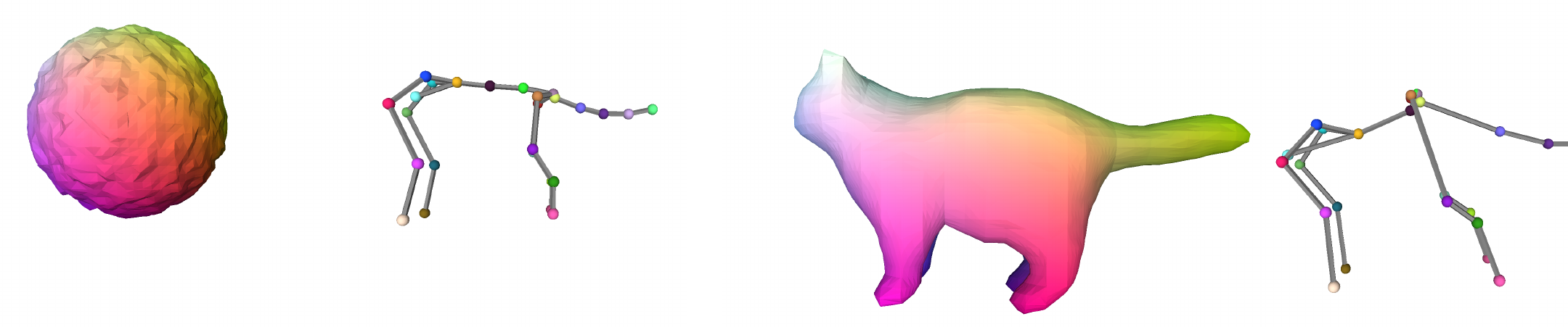}
    \includegraphics[width=\linewidth, trim={3cm 7.3cm 3cm 7.5cm},clip]{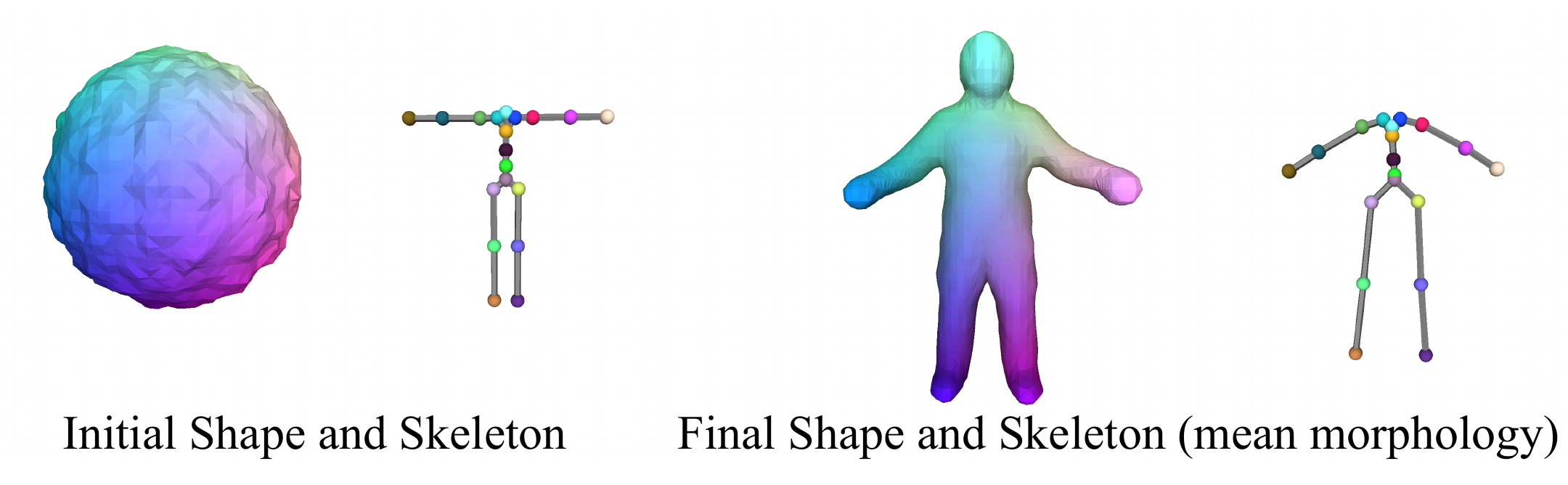}
    \caption{\textbf{Shape and skeleton optimization.} From top to bottom, we visualize the canonical shape and skeleton of our dog, cat, and human models. {\bf Left}: Canonical shape and skeleton before optimization. {\bf Right}: Canonical shape and skeleton after optimization.}
    \label{fig:skel-vis}
    \vspace{-10pt}
\end{figure}

\noindent{\bf Optimization Details}
We use AdamW to optimize the model for 36k iterations with 16384 rays per batch (taking around 24 hours on 8 RTX-3090 GPUs). We fist pre-train the background model with RGB, optical flow, and surface normal losses while ignoring foreground pixels. Then we combine background models with the object model for composite rendering optimization. The weights for the loss terms are tuned to have similar initial magnitude. %
The object root poses are initialized with single-image viewpoint networks trained for humans and quadruped animals following BANMo~\cite{yang2022banmo}.
For all categories, we start with the same shape (a unit sphere) and a known skeleton topology. Both the shape and the joint locations are specialized to the input dataset, as shown in Fig.~\ref{fig:skel-vis}.

\subsection{Reconstructing Humans}

\noindent\textbf{Dataset.} We combine existing human datasets, including AMA, MonoPerfCap, DAVIS, and BANMo to get 47 human videos with 6,382 images~\cite{Pont-Tuset_arXiv_2017, Xu:2018:MHP:3191713.3181973, vlasic2008articulated}. AMA contains multi-view videos, but we treat them as monocular videos and \emph{do not} use the time-synchronization or camera extrinsics. During preprocessing, we use PointRend~\cite{kirillov2020pointrend} to extract object segmentation, CSE~\cite{neverova2021discovering} for pixel features, VCN-robust~\cite{yang2019volumetric} for optical flow, and omnidata~\cite{eftekhar2021omnidata} for surface normal estimation. 

\noindent\textbf{Metrics.} We use AMA for evaluation since it contains ground-truth meshes and follow BANMo to compute both Chamfer distances and F-scores. Chamfer distance computes the average distance between the ground-truth and the estimated surface points by finding the nearest-neighbor matches, but it is sensitive to outliers. F-score at distance thresholds $d\in\{1\%,2\%,5\%\}$ of the human bounding box size~\cite{tatarchenko2019single} provides a more informative quantification of surface reconstruction error at different granularity.
To account for the unknown scale, we align the predicted mesh with the ground-truth mesh using their depth in the view space.

\noindent\textbf{Baselines.} On AMA, we compare with template-free BANMo~\cite{yang2022banmo} and model-based methods, including HuMoR~\cite{rempe2021humor} and ICON~\cite{xiu2022icon}. BANMo reconstructs an animatable 3D model from multiple monocular videos of the same instance, powered by differentiable rendering optimization. We optimize BANMo on the same dataset with the same amount of computation and iterations as ours. HuMoR is a temporal pose and shape predictor for humans. It performs test-time optimization on video sequences leveraging OpenPose keypoint detection and motion priors trained on large-scale human motion capture dataset. We run it on each video sequence, and processing 170 video frames takes around two hours on a machine with Titan-X GPU. ICON is the recent SOTA method on single view human reconstruction. It combines statistical human body models (SMPL) with implicit functions and is trained on 3D human scans with clothing. Notably, it performs test-time optimization to fit surface normal predictions to improve the pose accuracy and reconstruction fidelity. We run it per frame, and processing a 170 frame video takes around three hours on an RTX-3090 GPU.

\noindent\textbf{Results.} 
We show qualitative comparison with baselines in Fig.~\ref{fig:results-comp} top row, and quantitative results in Tab.~\ref{tab:quan-human}. On the handstand sequence, HuMoR works well for common poses but fails where the performer is not in an upright pose. ICON works generally well, but the hand distances appear not physically plausible (too short) from a novel viewpoint. BANMo reconstruction also failed to reconstruct the unnatural upside-down pose. In contrast, \ourmethod{} successfully reconstructs the handstand pose with plausible hand distances. On the samba sequence, HuMoR correctly predicts body poses, but fails to reconstruct the cloth and its deformation. ICON predicts a broken dress and distorted human looks from a novel viewpoint, possibly due to lack of diverse training data from dressed humans. When applied to 47 videos of different humans, BANMo fails to model the cloth correctly, possibly because a limited number of control points are not expressive enough to model the morphological variations over humans wearing different clothes. \ourmethod{} models between-shape variations using a conditional canonical model and successfully reconstructs cloth deformation using the soft deformation field.

Our quantitative results align with qualitative observations, where \ourmethod{} outperforms all baselines except being slightly worse than BANMo trained on single instances (S). However, \ourmethod{} trained on single instances (S) or multiple instances (M) outperforms BANMo trained in either fashion. In particular, BANMo results are notably worse when trained on multiple instances, indicating the difficulty in building category models. In contrast, \ourmethod{} become better when trained on multiple instances.

\subsection{Reconstructing Cats and Dogs}

\noindent\textbf{Dataset.} We collect 76 cat videos and 85 dog videos from Internet videos, as well as public data from BANMo. 
All the videos are casually-captured monocular videos. We extract video frames at 10 FPS, including 9,734 frames for cats and 11,657 frames for dogs. We perform the same pre-processing as human reconstruction. 

\noindent\textbf{Baselines.} We compare with BANMo and model-based BARC~\cite{BARC:2022}. BARC is the current SOTA for dog shape and pose estimation. It trains a feed-forward network using CG dog models and images with keypoint labels. The shape model is based on SMAL, which uses manual rigging and registration to fit 3D animal toys. We run BARC on individual images.

\noindent\textbf{Results.} We show qualitative results in Fig.~\ref{fig:results-comp} bottom row. For dog videos, we find that BARC worked well to predict coarse shapes and poses. However, the results are biased towards certain dog breeds. For instance, BARC predicts a long jaw when the dog has a short jaw (top row), and predicts round ears when the dog has sharp ears (bottom). BANMo was able to reconstruct a reasonable coarse shape, but failed to capture the fine details (e.g., the shape of the ear and the size of the head) with only 25 control points. In contrast, \ourmethod{} was able to faithfully capture the coarse shape and fine details of the dogs. Unlike dogs, cats have fewer variations in body shape and size, where we find that BANMo works well in most cases. However, for the body parts not visible from the reference viewpoint, BANMo often estimates a squashed shape, which may be caused by the entangled morphology and articulations. In contrast, \ourmethod{} accurately infers reasonable body parts and articulations even when they are not visible.

\subsection{Diagnostics}

\begin{figure}[h!]
    \centering
    \includegraphics[width=\linewidth, trim={7.5cm 9cm 7.5cm 8.5cm},clip]{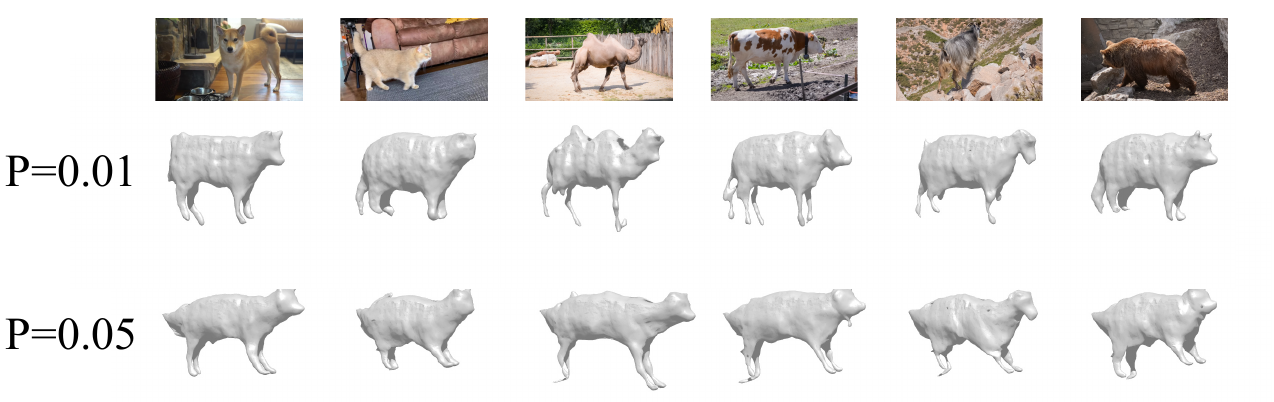}
    \caption{\textbf{Quadruped Category Reconstruction.} Using a smaller code swapping probability $P=0.01$ results in more faithful instance shape, but less smooth results. A larger $P$ produces smoother results, but some instance-specific features disappear.}
    \label{fig:quadruped}
\end{figure}

\noindent\textbf{Large Morphological Changes.} We reconstruct eight videos of different quadruped animals together to ``stress test'' our method. The dataset contains two dog videos, two cat videos, and one video for goat, bear, camel, and cow, respectively. The result is shown in Fig.~\ref{fig:quadruped}. 

\begin{figure}[h!]
    \centering
    \includegraphics[width=\linewidth, trim={6cm 8.8cm 6.5cm 9cm},clip]{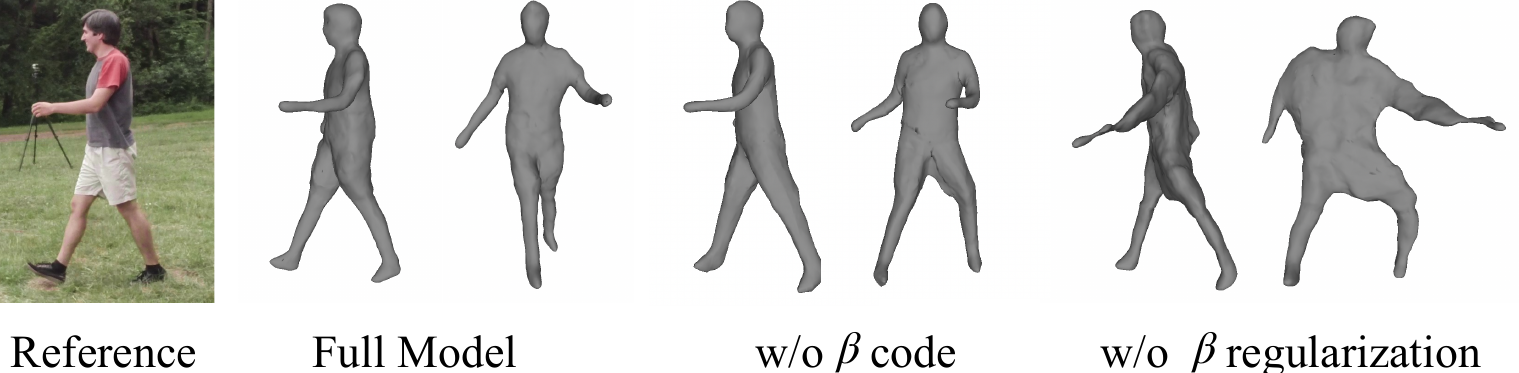}
    \caption{\textbf{Ablation study on morphology modeling}.}
    \label{fig:results-beta-reg}
\end{figure}

\noindent\textbf{Morphology Code $\boldsymbol{\beta}$} Removing morphology code ${\boldsymbol\beta}$ from the canonical field degrades it to a standard NeRF. We rerun the experiments and the results are shown in Tab.~\ref{tab:quan-human} as well as Fig.~\ref{fig:results-beta-reg}. Without the morphology code, our reconstructions are forced to share the same canonical shape, which as discussed in Sec~\ref{sec: morphology}, failed to handle fine deformations and topological changes (e.g., clothing), leading to worse results in all metrics.

\noindent\textbf{Morphology Code Regularization} To test the effectiveness of the morphology code regularization, we set $\mathcal{P}=0$ throughout the optimization. The results are shown in Tab.~\ref{tab:quan-human} and Fig.~\ref{fig:results-beta-reg}. Without regularization of the morphology code, the reconstructed shape may appear reasonable from the reference viewpoint, but severely distorted from a novel viewpoint. We posit the body parts that are not well-covered in the video are inherently difficult to infer. As a result, the shapes become degenerate without relying on priors from other videos in the dataset. Tab.~\ref{tab:quan-human} also shows that code swapping outperforms norm regularization~\cite{ghosh2020variational} ($||\boldsymbol\beta||^2_2$). We posit norm regularization forces codes to be similar, but does not constrain their output space, while code swapping encourages \emph{any} code to explain \emph{any} image in the data. In practice, we find that code swapping generates valid outputs when we interpolate between codes. 

\begin{figure}[h!]
    \centering
    \includegraphics[width=\linewidth, trim={6cm 9cm 6.5cm 9cm},clip]{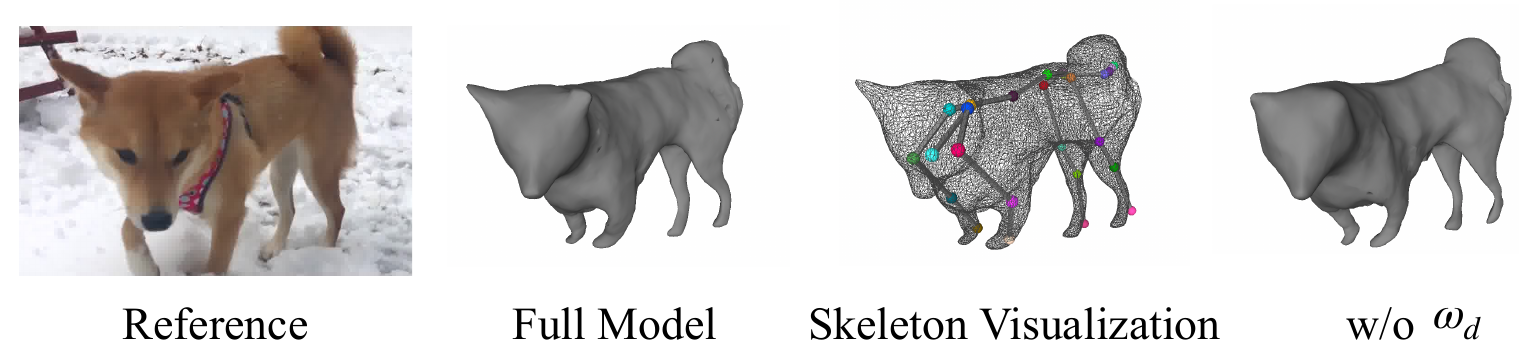}
    \caption{\textbf{Ablation study on soft deformation $\boldsymbol\omega_d$}.}
    \label{fig:aba-deform}
\end{figure}

\noindent\textbf{Soft Deformation.}
After removing the soft deformation field, \ourmethod{} fails to recover body parts that are not controlled by the skeleton (Fig.~\ref{fig:aba-deform}), such as the ears of the dog.

\noindent\textbf{Motion Transfer.} Given the category model with disentangled morphology and articulations, we can easily transfer an articulation in frame $t$ to other video by setting the articulation parameter to $\boldsymbol\theta_t$, while keeping the morphology parameter $\boldsymbol\beta$ the same. We show motion transfer across human in Fig.~\ref{fig:disentangle}. Please visit our website for video results of dogs, cats, and humans.

\begin{figure}[!h]
    \vspace{-2pt}
    \centering
    \includegraphics[width=\linewidth, trim={5.2cm 6.5cm 5.2cm 6.5cm},clip]{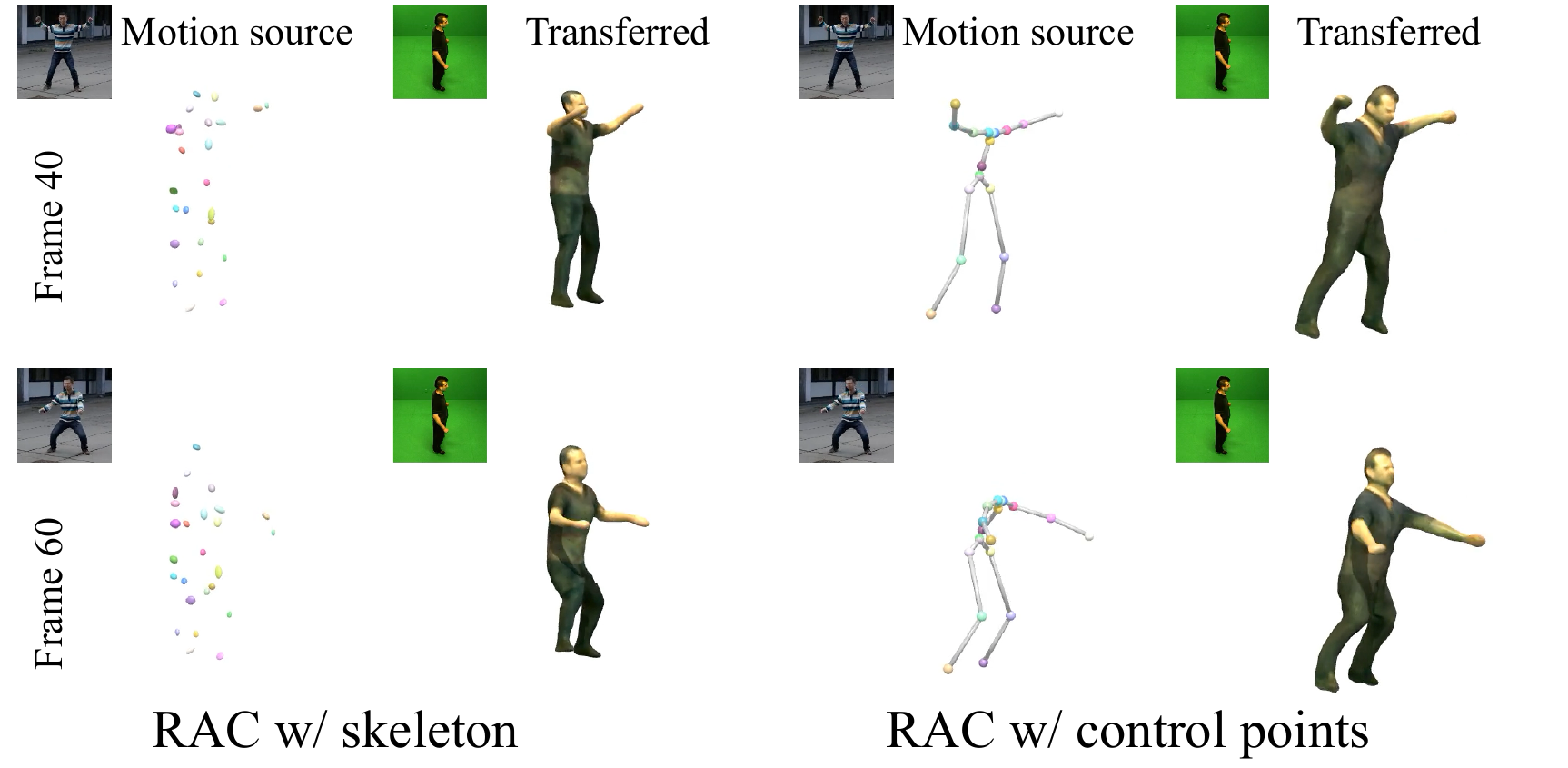}
    \caption{\textbf{Skeleton vs control-points for motion transfer}. \ourmethod{} with control points fails to maintain the body dimensions during motion, and produces squashy results when transferred to a new morphology. \ourmethod{} with skeleton disentangles motion from morphology, allowing for better motion transfer. }
    \label{fig:comparison-disentang}
\end{figure}

{\noindent \bf Skeleton vs Control Points.} Control point deformations are flexible but do not preserve body dimensions (e.g., a line segment can be stretched longer by its end points). As a result, body and limb dimensions can change, creating two problems: (1) articulated shapes look squashy from novel views, and (2) variations in body dimensions are entangled with control-point deformations. In contrast, skeleton deformation preserves body dimensions. It produces better results (Tab.~\ref{tab:quan-human}) and better motion re-targeting (Fig.~\ref{fig:comparison-disentang}).

{\noindent \bf Stretchable Bones} allow for control of bone dimensions after optimization. We show an example of a Dachshund (Source1) warped to a Heeler (Source2) by modifying bone dimensions while keeping the shape unchanged.

\begin{figure}[!h]
    \vspace{-10pt}
    \centering
    \includegraphics[width=\linewidth, trim={2cm 6.6cm 2cm 6cm},clip]{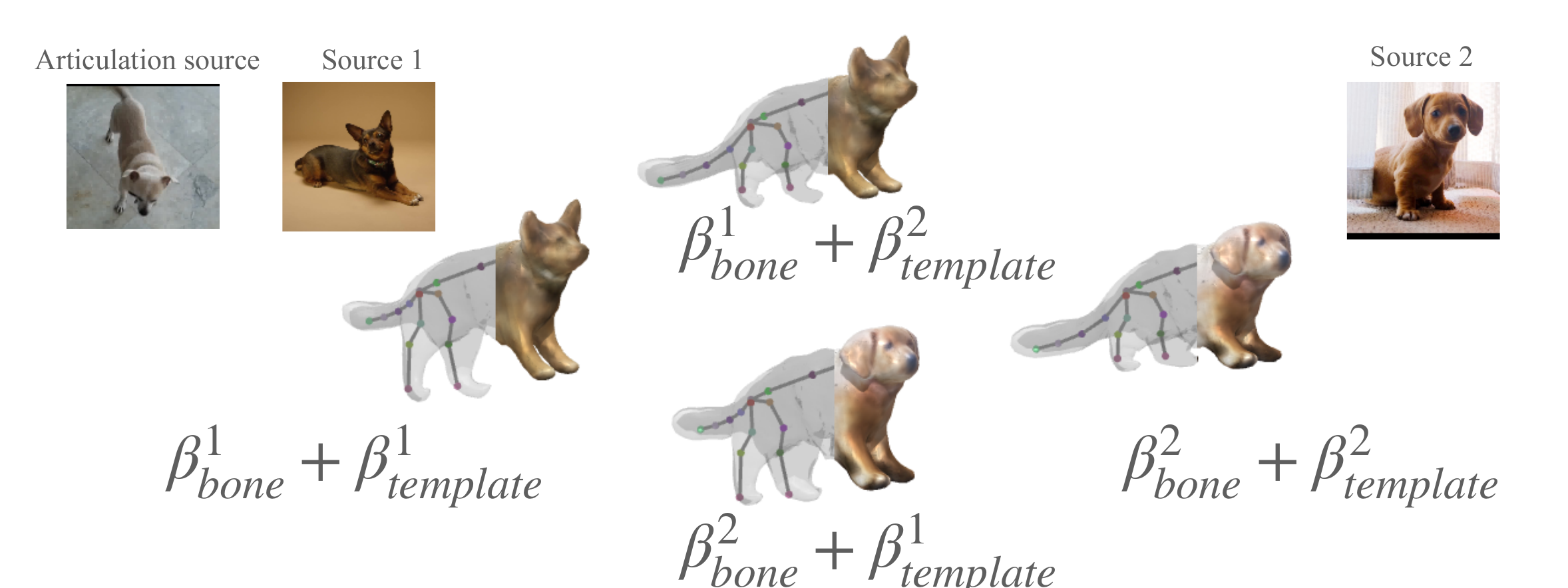}
    \caption{Disentanglement of bone dimensions and shape.}
    \label{fig:bone-shapel}
    \vspace{-10pt}
\end{figure}

\section{Discussion}

We have presented a scalable way of building animatable category models by learning from monocular videos. It disentangles morphology variations between instances and motion within an instance, allowing motion transfer over a category. \ourmethod{} reaches state-of-the-art reconstruction quality for cats, dogs, and humans in terms of mid-level reconstruction, but details are still missing (such as human hand and toes).
Similar to BANMo, \ourmethod{} requires rough viewpoint initialization. Although we have shown either a pre-trained viewpoint estimator or roughly annotated camera viewpoints (in the supplement) are sufficient, it would be interesting to study a more generic way to initialize viewpoints.
We also show that category-level skeleton improves motion reconstruction, and leave jointly inferring skeleton structure together with object shape for future work.

\noindent{\bf Acknowledgement} This work was supported by the Qualcomm Innovation Followship and the CMU Argo AI Center for Autonomous Vehicle Research.

\input{supp-text}

{\small
\bibliographystyle{ieee_fullname}
\bibliography{egbib}
}

\end{document}

%% file: supp-text.tex
\appendix

\section{Additional Details}
\noindent{\bf Shape Regularization.}
We apply eikonal regularization to force the norm of the first order derivative of signed distances $d$ to be close to 1,
\begin{equation}
\mathcal{L}_{\mathrm{eikonal}} = (\left\|\nabla \mathrm{\bf MLP}_{\mathrm{SDF}}({\bf X})\right\|-1)^2.
\end{equation}
Eikonal loss forces the reconstruction to be a valid surface and empirically improves the surface reconstruction quality. 

\noindent{\bf Pose, Deformation, and Appearance Smoothness.} 
We would like the time-varying articulated pose, deformation, and appearance codes $\{\boldsymbol\theta, \omega_d, \omega_a\}$ to vary smoothly within a video. To accomplish this, we make use of time-dependent positional embeddings (similar to ~\cite{yang2022banmo}):
\begin{equation}
    \omega_t^b = {\bf A}_i\mathcal{F}(t)
\end{equation}
where $\mathcal{F}(\cdot)$ is a 1D basis of sines and cosines with linearly-increasing frequencies at log-scale~\cite{tancik2020fourfeat}, and we learn separate weight matrices ${\bf A}_{i\in\{1\dots,M\}}$ for each video. %

\section{Category Outside DensePose} 

We test RAC in a scenario where there is no predefined DensePose features and skeleton.

\begin{figure}[!h]
    \centering
    \includegraphics[width=\linewidth, trim={9cm 9.5cm 10cm 8.5cm},clip]{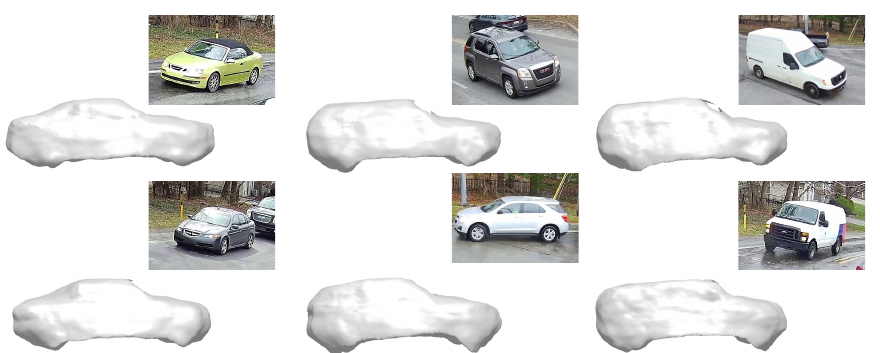}
    \caption{\textbf{Vehicle Category Reconstruction.} Our method is able to fuse videos of 365 vehicles with different appearance and shape into a category model. From left to right, we show reconstruction of sedans, SUVs, and vans.}
    \label{fig:car}
\end{figure}

\noindent\textbf{Vehicle Dataset.} 
We employ images from multiple 4K cameras \cite{Reddy_2022_CVPR} that overlook urban public spaces to analyze the flow of traffic vehicles. The data are captured for 3-second bursts every few minutes, and only images with notable changes are stored. We extracted 365 car videos from the dataset to build the category model. The dataset contains wide variation in vehicle categories like pickup trucks, construction vehicles etc on which traditional model based approached perform poorly. %

\noindent\textbf{Camera Pose Initialization.} As there is no DensePose model for cars, we took a two-stage approach to first coarsely register a few car videos with manual viewpoint annotation and then train a single-image viewpoint network to predict viewpoints for the rest of the videos. The camera viewpoints are roughly annotated for each frame (with around 30 degree rotation error). Annotation for a 100 frame video takes around 30 seconds. We found annotating 20 cars to be sufficient to train a viewpoint estimator that generalizes to other cars.

\noindent\textbf{Results.} We show the reconstruction results of car videos in Fig.~\ref{fig:car}. Please visit the website for more results.

\section{Evaluation on~\texttt{Pablo} Sequence} 
We compare with baselines on the \texttt{Pablo} sequence, which is part of the public MonoPerfCap~\cite{Xu:2018:MHP:3191713.3181973} dataset. Our method optimizes the \texttt{Pablo} sequence together with the rest of our 47 human videos. After differentiable rendering optimization, we extract meshes for the \texttt{Pablo} sequence and compare with the 3D ground-truth for evaluation.
 
\noindent\textbf{Metrics.} We follow the evaluation protocol of MonoClothCap~\cite{xiang2020monoclothcap} and compute the average point-to-surface distances in the clothing region. The clothing region (the T-shirt and shorts) is obtained by manual segmentation on the ground-truth surface mesh.

\begin{table}[t!]
    \caption{\textbf{Quantitative results on \texttt{Pablo} sequence.} 3D Chamfer distance (cm, $\downarrow$) is computed on the clothing region and averaged over all frames. MPCap uses a pre-scanned personalized template.}
    \small
    \centering
    \begin{tabular}{lcccccc}
	\toprule
	Method & MPCap* & MCCap & PiFuHD & T2S & \ourmethod{}\\
	\midrule
	Chamfer & 14.6 & 17.9 & 26.5 & 27.7 & 18.3\\
\bottomrule
\label{tab:quan-pablo}
\end{tabular}
\end{table}

\noindent\textbf{Results.}
We show quantitative comparisons in Tab.~\ref{tab:quan-pablo} and refer the reader to the qualitative results in Fig.~8 of the main draft. Our method outperforms PiFuHD~\cite{saito2020pifuhd}, Tex2Shape (T2S)~\cite{alldieck2019tex2shape}, both of which are single-view human shape predictors trained on 3D scans of humans. Our method does not use 3D data to train but performs test-time optimization on 47 human videos. Our method is slightly worse than MonoClothCap (MCCap)~\cite{xiang2020monoclothcap} that uses a parametric human body model (SMPL), and worse than MonoPerfCap (MPCap), which uses a prescanned template. Both parametric body model and personalized shape template provides a strong shape prior, while our method does not rely on any shape prior.

\section{Difference from prior works} We highlight the difference from previous work in Tab.~\ref{tab:difference}. In terms of shape modeling, {HyperNeRF}~\cite{park2021hypernerf} and {HumanNeRF}~\cite{weng_humannerf_2022_cvpr} reconstruct a \emph{single} scene or instance, while \ourmethod{} learns a space of category shapes. For skeleton modeling, {CASA}~\cite{wu2022casa} is optimized \emph{per-instance}, while \ourmethod{} learns a shared space over a category of skeletons (with different bone lengths). For background modeling, {NeRF++}~\cite{zhang2020nerf++} assumes a static scene and does not use background to help object segmentation and reconstruction. {NerFace}~\cite{gafni2021dynamic} treats background as a static image, while we represent the background as a NeRF, which generalizes to videos captured by a moving camera.
 
\begin{table}[h!]
\vspace{-5pt}
    \caption{\textbf{Difference between prior works and \ourmethod{}.}}
    \vspace{-10pt}
    \centering
    \resizebox{\columnwidth}{!}{%
    \begin{tabular}{lcccc}
	\toprule
Method & Shape & Motion & Background & 3D Data/Pose\\
\midrule
NeRF++ & N.A. & N.A. & NeRF & No\\
NeRFace & Instance & Conditional & Image & No\\
HyperNeRF  & Instance &Fields+Conditional & N.A. & No\\
BANMo  & Instance &Control Points & N.A. & No\\
CASA & Instance & Instance Skeleton & N.A. & Yes\\
HumanNeRF  & Instance &Instance Skeleton & N.A. & Yes\\
\ourmethod{}   & Category & Category Skeleton & NeRF & No\\
\bottomrule
\label{tab:difference}
\end{tabular}
}
\vspace{-15pt}
\end{table}